# Accurate pedestrian localization in overhead depth images via Height-Augmented HOG

**Werner Kroneman, Alessandro Corbetta, Federico Toschi**

Eindhoven University of Technology
Department of Applied Physics
Postbus 513, 5600 MB, Eindhoven, The Netherlands
w.kroneman@tue.nl; a.corbetta@tue.nl; f.toschi@tue.nl

***Abstract*** – We tackle the challenge of reliably and automatically localizing pedestrians in real-life conditions through overhead depth imaging at unprecedented high-density conditions. Leveraging upon a combination of Histogram of Oriented Gradients-like feature descriptors, neural networks, data augmentation and custom data annotation strategies, this work contributes a robust and scalable machine learning-based localization algorithm, which delivers near-human localization performance in real-time, even with local pedestrian density of about 3 ped/m$^2$, a case in which most state-of-the art algorithms degrade significantly in performance.

## 1. Introduction

In order to design more pedestrian-friendly environments or improve the evacuation performance of buildings, developing reliable quantitative models of crowd behavior is paramount. The formulation and validation of models demands the acquisition of accurate, real-life, experimental data, e.g. in the form of individual trajectories. Due to the variability of pedestrian behavior, large data volumes are required to robustly isolate average behaviors and obtain resolved statistics of fluctuations and rare events [1,2].

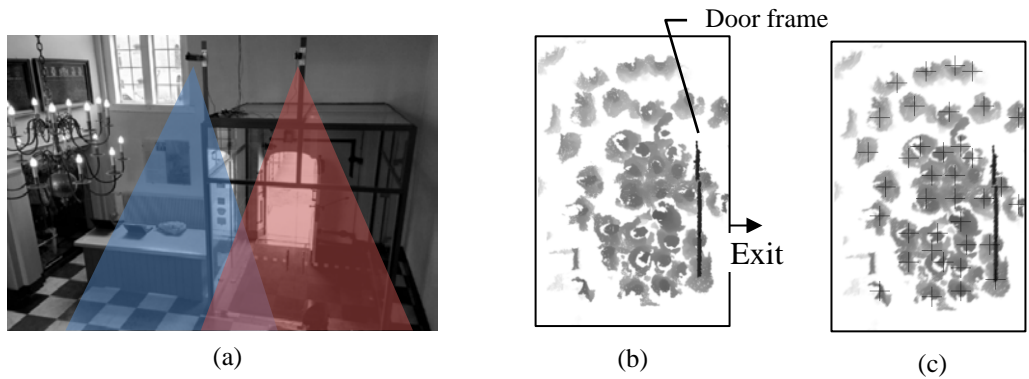

Figure 1: (a) Depth-based tracking system composed of two overhead depth sensors (Microsoft Kinect®) [3] deployed at the entrance of the Pesthuis at the Naturalis Biodiversity Center in Leiden (photo by AC). (b) Sample of a depth image. The depth field is colorized such that points of lower depth, i.e. closer to the sensor and higher above the ground, are in a darker shade. This depth image is taken a moment before a school group leaves through the door. As can be seen in (b), extremely high crowd densities occur at this site, and individual localization is difficult even for a human expert. Pedestrian positions are reported in (c) as output by the HA-HOG method introduced in this paper. This method delivers near-human-quality localization in these conditions, considerably improving upon the state of the art.

However, for the strong variability in pedestrian shapes and environments, achieving quantity and quality of measurements while operating in real-life conditions is a substantial technical challenge. Moreover, privacy regulations for public spaces dictate that recorded images of pedestrians must be anonymous.

Since 2013, overhead depth sensing has shown remarkable capabilities for real-life pedestrian localization, and has been employed in a number of recent experiments [2, 4, 5, 6]. Depth images map the distance between each point and the camera plane. Their usage, in place of color images, allows one to bypass difficulties related to varying light and color as the depth maps directly capture the 3-dimensional shape of pedestrians. Besides, as can be seen in Figure 1b, depth images also contain insufficient information to personally identify pedestrians, which ensures anonymity. Typically, sensors (e.g. Microsoft Kinect® [3]) are placed overhead, facing downwards, to avoid mutual pedestrian occlusions in the captured signals. Multiple cameras can be combined in grid-like configurations to cover larger areas as in Figure 1 or in [4, 7].

Individual tracking is typically implemented as a two-step process: 1. localizing pedestrians in consecutive depth images; 2. linking these detections across time into tracks. So far, the localization step has been a challenge. Especially in dense crowds where pedestrians walk or stand closely together, the most commonly-used method, clustering [6], exhibits a substantial performance degradation, producing errors such as those depicted in Figure 2. Besides, achieving high-frequency localization is paramount to minimize tracking error, so high algorithmic and implementation efficiency is a must.

In this contribution, we present a robust and scalable localization method based on a Histogram-of-Oriented-Gradients-like [8] feature descriptor and a neural network classifier, which remains accurate even at high density. This paper is structured as follows: in Section 2, we elaborate on bottlenecks in state-of-the-art techniques. In Section 3, we introduce our method whose performance is discussed in Section 4. A final discussion concludes the paper.

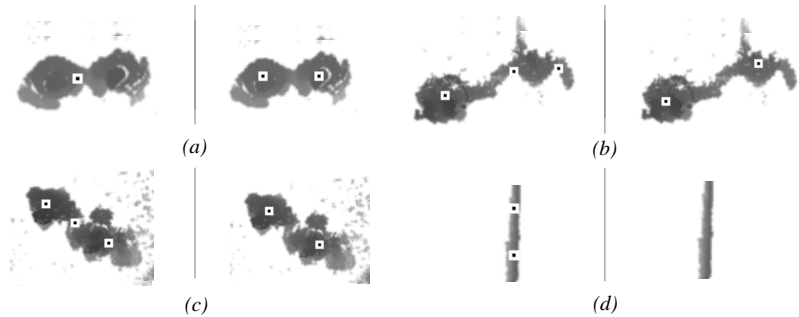

Figure 2: (left panels, a-f) Common errors made by overhead depth-map-based localization algorithms (here considering the clustering method in [6]), (right panels, a-f) ground truth. Specifically: (a) two pedestrians detected as one, (b) a single pedestrian detected multiple times, (c) a hand detected as a pedestrian and (d) a wall confused for a pedestrian. Such errors are made because the clustering algorithm has almost no notion of pedestrian shape, and therefore cannot easily distinguish pedestrians when standing so close together, nor tell the difference between pedestrians and other objects.

## 2. Related work

To the best of our knowledge, the only techniques employed to localize pedestrians in real-life depth-based crowd dynamics investigations are clustering-based (e.g. [2, 4, 5, 6]). Typically, these techniques involve, first, foreground/background separation based on a height threshold. Then, the point clusters that are left in the foreground are interpreted as persons. At low density, such a pedestrian-cluster association is very effective as wide mutual distances ensure that never two people are clustered together. Besides, clustering is a simple method, with most clustering algorithms only requiring a small number of parameters. However, the accuracy drops severely at higher pedestrian densities, exhibiting problems as shown in

Figure 2. Moreover, clustering becomes computationally expensive as the size or complexity of the depth map increases, as stated in [9].

Awareness of pedestrian shape is necessary to bypass the bottlenecks of clustering (Figure 2), and thus segment pedestrians standing close together. However, countless exceptions are encountered in real-life data, which result from variations in pedestrian shape or imperfections in the depth image. Hence, writing a shape-aware algorithm which achieves sufficiently high localization performance is extremely difficult, with constant requests for the treatment of newly-discovered exceptional cases.

One possible solution to this problem, recently exploding in popularity, is the use of deep learning models [10]. These models are trained end-to-end by annotated examples, and can then generalize towards cases they were never presented with. This, however, requires a large set of depth images manually-annotated with pedestrian positions, incredibly labor-intensive to obtain. To bypass this problem, in [11], a convolutional neural network is trained end-to-end employing synthetically-generated training data. Although the method is reported to have better performance than clustering-based approaches, precision and recall values still dropped below 0.95 at higher pedestrian densities, which is lower than achieved with our method. Besides, synthetically generating realistic high-density depth images is likely a highly non-trivial task.

In the next section, we present a technique inspired by "classic" machine learning approaches which combines learning by examples and traditional localization algorithms. This ensures both robustness against exceptions and low cost of training.

## 3. HA-HOG localization

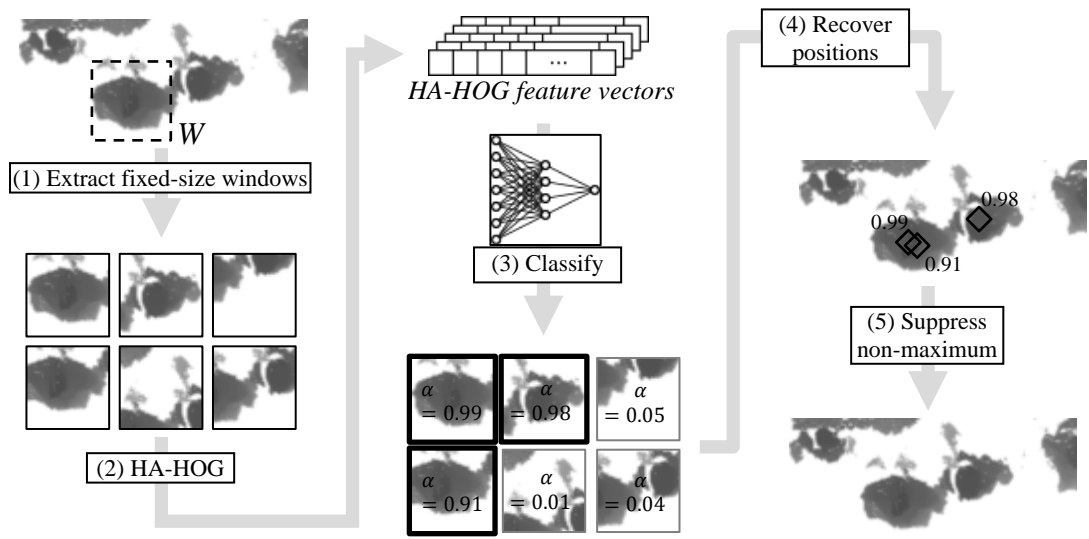

Figure 3: Outline of the localization process. (1) Slide a window with a fixed size slightly larger than a pedestrian over the depth map and process the content separately. (Only a small number are shown here for illustration purposes.) (2) Compute the HA-HOG of each of these samples (3) and present these feature vectors to a neural network, which assigns a score to each sample based on whether it is centered on a pedestrian or not (the score is approximately 1 or 0 in each case, respectively). By applying a threshold, translate this score to a binary "positive" or "negative" decision. (4) Next, associate the positively-classified samples with positions in the depth map. (5) Since multiple windows can be roughly centered on the same pedestrian, this pedestrian may be detected multiple times. Therefore, apply non-maximum suppression to remove duplicate detections. Finally, return the resulting set of positions as the set of detected pedestrian locations within the depth image.

In order to localize pedestrians in a depth image, we slide a fixed-size window throughout the image and classify its content depending on whether it is centered on a pedestrian or not. The process is illustrated in Figure 3 and explained in the next sections. Specifically, in Section 3.1 we present the HA-HOG feature descriptor and in Section 3.2 we explain its usage for localization. Training and implementation details are provided in Section 3.3.

### 3.1. HA-HOG Feature description computation

"Image descriptors" are methods that take an image and return a summarized representation, isolating information relevant to the problem. Such simpler representations require less complex statistical models (and therefore training data) when it comes to content classification. In our method, we adapt a simplified version of "Histogram of Oriented Gradients" (HOG), originally introduced by Dalal and Triggs [8] for the detection of humans in color images. Let $W$ be the content of a window sliding through a depth image. The HOG features of $W$ are computed as follows:

1. compute the gradient field $\vec{G} = \nabla W$ (e.g. using the second order finite differences method);
2. compute the polar representation of $\vec{G}$, $(G_r, G_\varphi)$, such that:
$$\vec{G} = (G_x, G_y) = G_r \cdot \begin{pmatrix} \cos G_\varphi \\ \sin G_\varphi \end{pmatrix} \text{ and } 0 < G_\varphi < 2\pi;$$
3. partition $\vec{G}$ into a grid of small, non-overlapping cells. In our case, these cells are square and five or six pixels wide. In each cell, make a histogram of the gradient vectors, binned using $G_\varphi$ and weighted by $G_r$. Finally, the histogram is normalized to have unit L2-norm. In formulas, the HOG feature of a cell reads:
$$HOG_{Cell} = \frac{Histogram(G_\varphi, weights = G_r)}{|Histogram(G_\varphi, weights = G_r)|};$$
4. concatenate the HOG of each cell in the window to yield the HOG descriptor of the window W.

$$HOG(W) = [HOG_{Cell}(c_1), HOG_{Cell}(c_2), \dots, HOG_{Cell}(c_n)].$$

To further increase the classification accuracy (see Section 3.2), we integrate height information in the feature descriptor, the result of which we call Height-Augmented HOG (HA-HOG), i.e. algorithmically

5. compute a histogram of the depth values over $W$, which we then concatenate with the HOG:

$$HA\text{-}HOG(W) = [HOG(W), Histogram(heights \text{ in } W)].$$

Step 5 is aimed at allowing the classifier to avoid the misclassification of objects that are similar in shape to pedestrians, but differ significantly in height distribution.

### 3.2. Classification and localization

After computing the HA-HOG of each image window $W$, the resulting feature vectors are presented to a multilayer perceptron ([12], see [13] for an example of HOG used in combination with a neural network classifier). The neural network is trained through back-propagation to return a value α, with $0 \leq \alpha \leq 1$, near 1 for samples that contain a pedestrian in the middle ("positive"), and near 0 for ones that either do not contain a pedestrian or are not centered on one ("negative"). All windows for which $\alpha \geq t$ $(t \cong 0.9)$ are retained and their center is considered as a candidate for a pedestrian location.

Since the window is moved in strides which are much smaller than the size of a person, multiple consecutive windows may be roughly centered on a pedestrian and be classified as positive. This results in a pedestrian detected multiple times. To prevent this, non-maximum suppression is applied on the resulting set of positions: for pairs of locations that lie closer to each other than a certain threshold, the location with a lower α is discarded. This process is repeated until no such pairs of locations exist. The resulting set of locations is then interpreted as the locations of pedestrians within the image.

### 3.3. Training and implementation details

Creating a sufficiently large set of annotated training examples by manually annotating thousands of examples is labor-intensive human work. To bypass this bottleneck, we rely on image augmentation: in our case applying random mutations to image samples such as rotations of 90 degrees and noise insertion, effectively yielding a much larger set of training examples, also done in [14, 15]. Moreover, we also developed a tool which runs the algorithm on a random image and presents the result to a human expert, who can then quickly extract new training examples based on a visual comparison of the set of detected pedestrian locations and the original depth map, thereby targeting weaknesses in the training set. This is essentially a form of hard-mining or bootstrapping [16].

In order to reduce the computational cost of the HOG computation, setting the window stride to a multiple of the HOG cell size will cause many windows to have a large number of overlapping HOG cells, which will only have to be computed once. In our implementation, HOG cells are computed at once over the whole image, and then concatenated in overlapping blocks corresponding to the samples that would be taken by a sliding window on the original depth map.

We compute the HOG features of the image in TensorFlow [17] so as to be able to use the GPU. We implement, train and evaluate the neural network itself using the Keras machine learning framework [18]. Using the GPU capabilities of TensorFlow and Keras, the method can detect pedestrians at over 20 fps on a GT-1030 graphics card, even on depth images as large as those collected during the Glow 2017 campaign (see Figure 4 and [7]).

## 4. Performance

In order to assess the quality of the localization provided by HA-HOG, we manually annotated a set of depth images (disjoint from the training set), and compared our method against classic HOG and an implementation of the clustering-based method (cf. Section 2) used in [1, 2, 4].

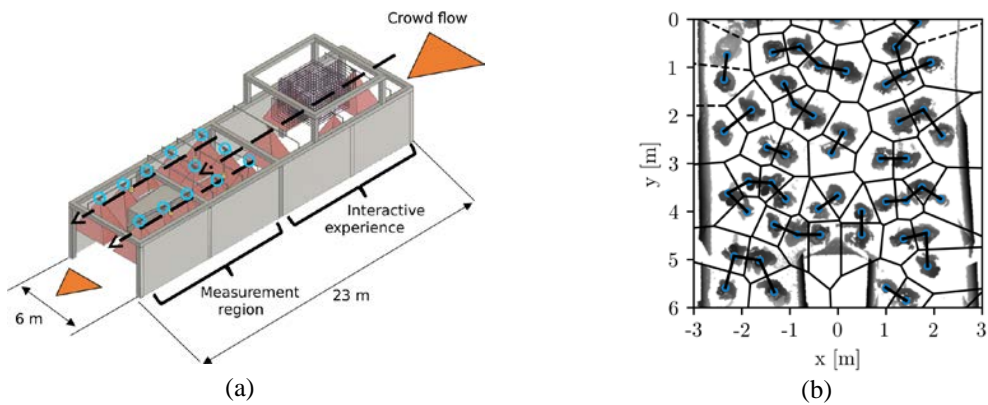

(a) (b)

Figure 4: (a) Diagram of the "Moving Light" exhibit at the Glow 2017 light festival in Eindhoven [Glow 2017]. It features a long, mostly continuous stream of people who walk through a large, grid-like set-up of twelve depth cameras (Microsoft Kinect®), circled in blue. Our GPU implementation of HA-HOG can localize pedestrians in this data with high accuracy while still running at over 20 fps. (b) Example of a depth map from the dataset collected at the exhibit annotated with pedestrian positions. A Voronoi tessellation of these annotations is provided and nearest-neighbor relations are displayed as well. All false-positives, false-negatives and true positives are associated with a ground truth position based on the Voronoi cell they fall into, and are thus assigned a nearest-neighbor distance. See Section 4.2 for details.

## 4.1. Dataset description

We benchmark the algorithm against data from 2 real-life locations:

- The Glow festival in Eindhoven during November in 2017 [7], see Figure 4, featuring a set of twelve cameras, yielding a challenge for computational performance. For instance, the clustering method becomes too computationally expensive to use on images of this size without severely compromising localization accuracy. This data features many people walking close together in small groups in which individuals can be hard to tell apart;

- The entrance of the Pesthuis of the Naturalis Biodiversity Center in Leiden between January and May 2018, see Figure 1, featuring extremely high crowd densities and complex scenes, a real challenge for localization even for human experts;

From both datasets, about 80 full-scale frames with high crowd densities were extracted, and annotated by hand (containing a total of, respectively, 1059 and 600 annotations). No training samples are taken from these frames to ensure that the results do not merely result from over-fitting.

## 4.2. Density-dependent performance analysis

We localize pedestrians in the hand-annotated test frames (see Section 4.1) using HA-HOG, classic HOG and complete linkage clustering, the result of which is then compared to hand-annotated positions, from which we identify true positive (TP), false positive (FP) and false negative (FN) localizations. Let $r_{nn}$ be the distance between each manual annotation and its nearest neighbor. We use the density $\rho_{nn}$, evaluated as $\rho_{nn} = 1/(\pi \cdot r_{nn}^2)$, as an indication of localization difficult. In this case, high density values directly imply close vicinity of the nearest neighbor (see Figure 4b). We bin each TP, FP and FN by $\rho_{nn}$ (taken from the closest manual annotation). Then, for each $\rho_{nn}$ bin, let #TP be the count of true positives in that bin (and analogously for #FP and #FN). We evaluate localization performance in terms of density-dependent F-score, defined as 2·(Precision·Recall)/(Precision+Recall), where precision and recall are defined as (#TP)/(#TP+#FP) and recall (#TP)/(#TP+#FN) respectively.

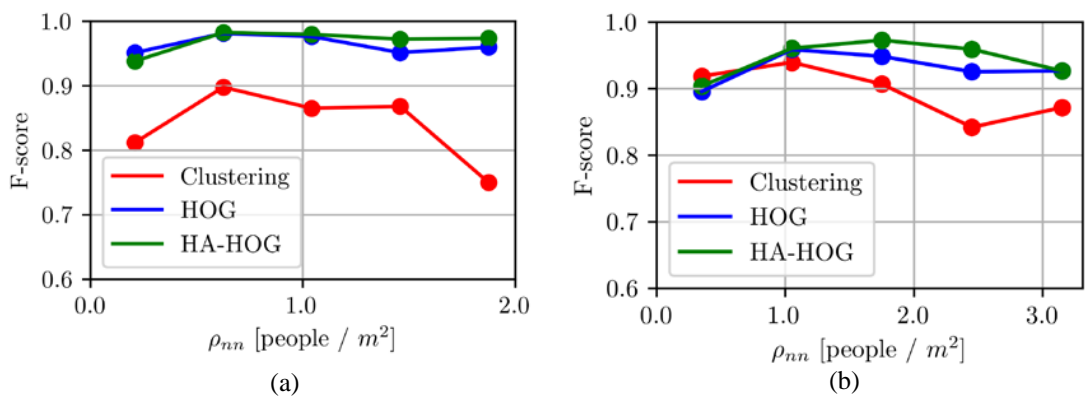

Figure 5: F-score of the localization methods as a function of the distance between closest pairs of pedestrians, when performing on (a) the Glow 2017 dataset and (b) on the Naturalis dataset. Both datasets are presented in Section 4.1. Nearest-neighbor distance is plotted on the horizontal axis with distance decreasing, and difficulty increasing, from left to right. For definitions and a discussion of the results, see Section 4.2.

On the Glow dataset (Figure 5a), HA-HOG performs comparably to HOG, and significantly out-performs clustering at all densities, especially around $\rho_{nn} = 2$ ped/m². On the Naturalis dataset, on the other hand, HA-HOG shows a significant performance gain over both classic HOG and clustering especially at particularly challenging conditions occurring for $\rho_{nn} > 2$ ped/m².

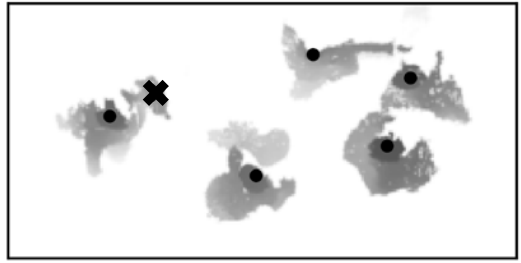

Figure 6: Depth map annotated with pedestrian positions using HA-HOG, with a dot for true positives, and a cross for false positives according to a comparison with manually-annotated positions. Note the false positive on the left: from looking at this depth image, it is difficult to tell whether this is really a false positive. The height and size roughly correspond to a small child, so the output of HA-HOG may be justified in this case despite it counting against the F-score.

At the lowest density bin, one may notice the F-score of HA-HOG dropping below 0.9. Part of this can be attributed to statistical noise, as the number of ground-truth annotations for this bin is only around 50. Besides we found that cases where even human experts fail to find agreement in annotation (cf. Figure 6) are generally associated with positive localization by HA-HOG, whilst being generally ignored by clustering. This causes a higher density-dependent F-score for the clustering method.

## 5. Discussion

In this paper, we presented HA-HOG, a technique for localizing pedestrians in overhead depth images robust even in high-density cases. The technique extends the classic Histogram of Oriented Gradients feature descriptor to include information about height distribution easily accessible from depth images.
The methods leverages awareness of shapes to achieve localization, which allows to outperform current approaches based on clustering, especially in high-density conditions. Furthermore, being trained by examples, the method remains robust against exceptions in shape, ubiquitous in real-life data. As annotated example are used just to train a final classifier, rather than the whole feature extractor (as done in deep learning-based approaches), they can be produced with limited human labor (few hours).

To evaluate performance we employed two datasets acquired in two real-life experiments in both of which our method showed near-human performance, with F-score above 95% at around 2.5 ped/m$^2$. Finally, unlike clustering, HA-HOG can be efficiently parallelized, e.g. on a GPU, with the computational cost scaling only linearly with the number of pixels.

## Acknowledgements

This work is part of the JSTP research programme "Vision driven visitor behavior analysis and crowd management" with project number 341-10-001, which is financed by the Netherlands Organization for Scientific Research (NWO). We acknowledge the support of Naturalis Biodiversity Center for hosting our measurement setup. We thank all contributors of the Moving Light experiment (in Figure 4a) for which we acknowledge the support of NWO, 4TU, Glow, Philips Lighting and Student Hotel Eindhoven. Figure 4a is made in collaboration with Bert Maas.